\documentclass[letterpaper]{article} 
\usepackage{aaai2026}  
\usepackage{times}  
\usepackage{helvet}  
\usepackage{courier}  
\usepackage[hyphens]{url}  
\usepackage{graphicx} 
\usepackage{natbib}  
\usepackage{caption} 
\usepackage{algorithm}
\usepackage{algorithmic}
\usepackage{amsmath}
\usepackage{amssymb}
\usepackage{booktabs}
\usepackage{pifont}
\usepackage{makecell}  
\usepackage{multirow}    
\usepackage{graphicx}    
\usepackage[table,xcdraw]{xcolor} 

\usepackage{newfloat}
\usepackage{listings}
\DeclareCaptionStyle{ruled}{labelfont=normalfont,labelsep=colon,strut=off} 
\floatstyle{ruled}
\newfloat{listing}{tb}{lst}{}
\floatname{listing}{Listing}
\title{Diffusion Once and Done: Degradation-Aware LoRA for Efficient All-in-One Image Restoration}
\author{
    Ni Tang\textsuperscript{\rm 1},
    Xiaotong Luo\textsuperscript{\rm 1},
    Zihan Cheng\textsuperscript{\rm 1},
    Liangtai Zhou\textsuperscript{\rm 1},\\
    Dongxiao Zhang\textsuperscript{\rm 2},
    Yanyun Qu\textsuperscript{\rm 1}
}
\affiliations{
    \textsuperscript{\rm 1}School of Informatics, Xiamen University, Xiamen, China\\
    \textsuperscript{\rm 2}School of Science, Jimei University, Xiamen, China\\

}

\usepackage{bibentry}

\begin{document}

\maketitle

\begin{abstract}
Diffusion models have revealed powerful potential in all-in-one image restoration (AiOIR), which is talented in generating abundant texture details. The existing AiOIR methods either retrain a diffusion model or fine-tune the pretrained diffusion model with extra conditional guidance.
However, they often suffer from high inference costs and limited adaptability to diverse degradation types. In this paper, we propose an efficient AiOIR method, Diffusion Once and Done (DOD), which aims to achieve superior restoration performance with only one-step sampling of Stable Diffusion (SD) models. Specifically, multi-degradation feature modulation is first introduced to capture different degradation prompts with a pretrained diffusion model. Then, parameter-efficient conditional low-rank adaptation integrates the prompts to enable the fine-tuning of the SD model for adapting to different degradation types. Besides, a high-fidelity detail enhancement module is integrated into the decoder of SD to improve structural and textural details. Experiments demonstrate that our method outperforms existing diffusion-based restoration approaches in both visual quality and inference efficiency.
\end{abstract}



\section{Introduction}

All-in-One Image Restoration (AiOIR) aims to handle multiple degradation tasks within a unified framework, thereby enhancing the robustness and deployment efficiency of image restoration models. The existing methods typically improve model adaptability through scalable network architectures \cite{wang2024tanet, zeng2025vision, tian2025degradation}, multi-task learning strategies \cite{wu2024harmony}, and integrating task-specific and task-agnostic priors \cite{yan2025textual, luo2023controlling}. However, most of these methods primarily focus on optimizing distortion-based metrics (e.g., PSNR and SSIM), while ignoring the equally important perceptual quality (e.g., LPIPS and MANIQA).

Recently, diffusion-based AiOIR methods \cite{xiong2025da2diff, zheng2024selective, jiang2024autodir, ai2024multimodal} have gained increasing attention, owing to the strong potential in modeling complex natural image distributions. It favors more realistic outputs with superior perceptual quality. Existing diffusion-based approaches primarily fall into two categories:
1) Train diffusion models from scratch with the specific dataset \cite{xiong2025da2diff, zheng2024selective}. While offering the generative ability for customized tasks, they demand significant training expense and large-scale datasets to obtain a robust model.
2) Fine-tune the pre-trained diffusion models (e.g., Stable Diffusion \cite{rombach2022high}) \cite{jiang2024autodir, ai2024multimodal}. 
By means of the powerful diffusion prior in structure and texture generation, these approaches typically promote image restoration effects via adding conditional guidance or extra enhancement modules, while they still struggle to adapt to different degradation types. 
Note that these two types of methods rely on multi-step sampling, resulting in low inference efficiency that hinders practical deployment. Hence, it is desirable to propose an efficient diffusion-based method for robust AiOIR, which can reduce burdensome inference costs.

\begin{figure}[t] 
    \centering
    \includegraphics[width=1\linewidth]{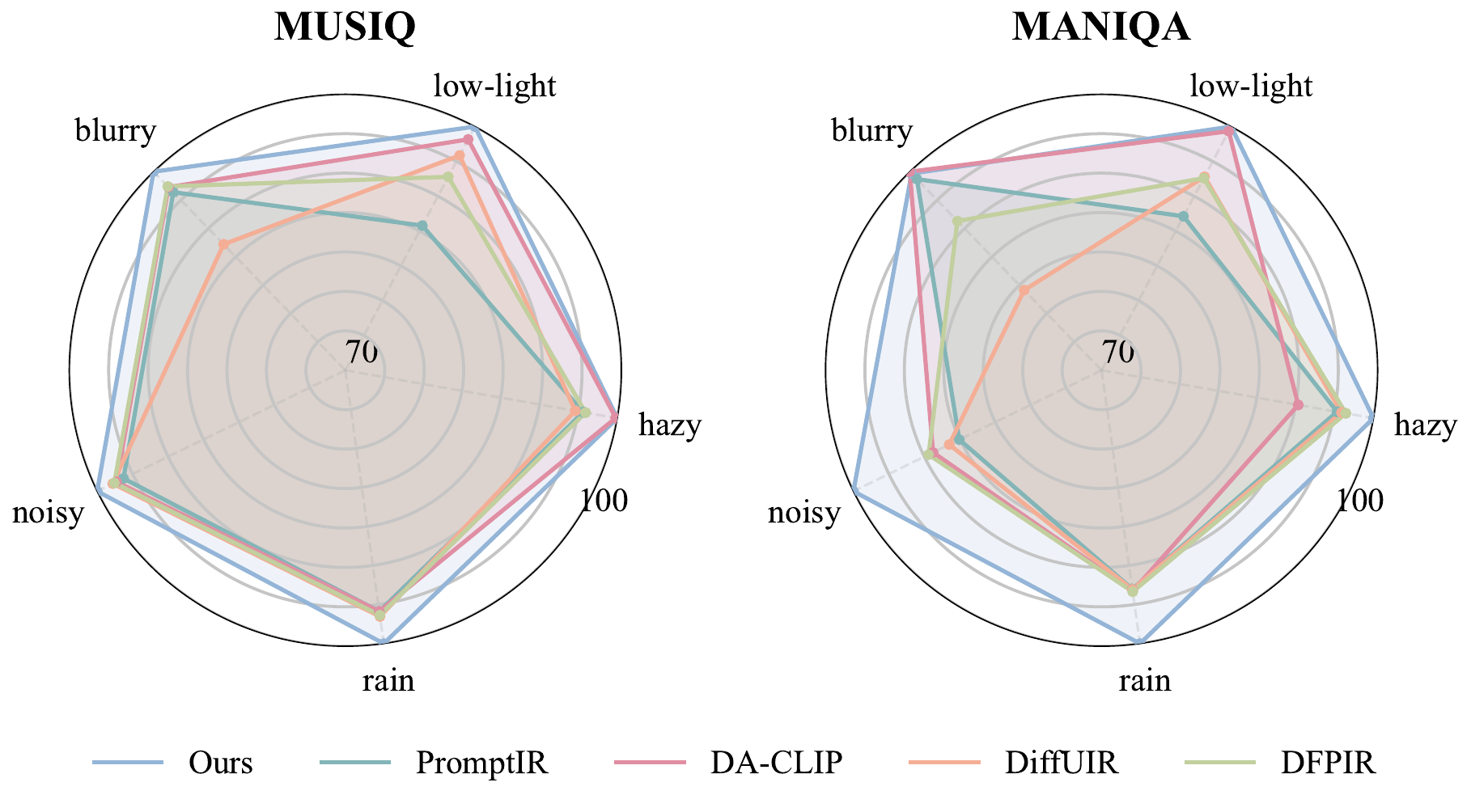} 
    \caption{
    Comparison on five restoration tasks using MANIQA and MUSIQ metrics shows that our method consistently outperforms others, demonstrating its effectiveness across diverse degradation conditions.
    }
  \label{fig:rader_comparison}
\end{figure}

In this paper, we aim to design an efficient one-step diffusion method based on the pretrained Stable Diffusion (SD) for AiOIR, which can not only accelerate the inference but also maintain a superior restoration effect. However, it would face several challenges: 
1) how to effectively characterize diverse image degradation patterns to optimize and guide diffusion models for robust image restoration;
2) how to perform model fine-tuning while preserving the original parameter priors of the SD;
3) how to compensate for the high-quality detail loss inherent to one-step sampling.

To address the above problems, we propose a novel framework, Diffusion Once and Done (DOD), which is designed as a degradation-aware one-step diffusion solution for efficient and high-quality image restoration. Inspired by the efficient fine-tuning capability of Low-Rank Adaptation (LoRA) \cite{hu2022lora}, we design a degradation-aware LoRA strategy that dynamically modulates the model’s feature space under varying degradation conditions, enabling effective modeling of diverse degradation patterns.

Specifically, our DOD includes Multi-degradation Feature Modulation (MFM), Parameter-efficient conditional LoRA (PLA), and High-fidelity Detail Enhancement (HDE). 
MFM aims to extract separate degradation information from intermediate features of the pretrained DDPM \cite{ho2020denoising}, which can effectively distinguish different degraded types.
PLA integrates the degradation prompts with LoRA for feature space refinement, enabling a single set of parameters to adapt to different degradation types.
In HDE, we design the detail enhancement module to 
promote the texture and detail reconstruction.
The whole method is optimized in a two-stage way to achieve fast and high-quality restoration.
The two-stage training strategy ensures both computational efficiency through parameter-efficient tuning and performance superiority via degradation-aware adaptation and detail refinement.
Extensive experiments demonstrate that our DOD achieves superior performance across multiple AiOIR benchmarks (as illustrated in Fig. \ref{fig:rader_comparison}). Notably, it reduces the sampling steps from multiple iterations to just one step, significantly improving inference efficiency. Meanwhile, with minimal parameter fine-tuning, our approach attains exceptional restoration quality, showcasing its strong practical value and deployment potential for real-world image restoration applications.

In summary, our contributions are as follows:
\begin{itemize}  
    \item We propose a one-step diffusion model for AiOIR that leverages abundant priors from SD, achieving high-quality restoration with significantly reduced inference time and trainable parameters.

    \item We introduce a degradation-aware LoRA strategy, which adaptively modulates the feature space based on degradation cues extracted from a pre-trained diffusion model, enabling effective removal of diverse degradations.

    \item We further incorporate a detail enhancement module in the decoder of SD to improve the restoration of textures and structural details. 
    Besides, we adopt a two-stage optimization strategy to enable stable training.

    \item Extensive experiments show that our DOD achieves the best perceptual metrics with higher reconstruction accuracy on multiple image restoration tasks.
\end{itemize}

\section{Related Works}
\subsection{All-in-one Image Restoration}
Existing AiOIR methods typically improve generalization by architectural design and prompt mechanisms. For example, AirNet \cite{li2022all} introduces a degradation encoder, TransWeather \cite{valanarasu2022transweather} uses a query-based Transformer, PromptIR \cite{potlapalli2023promptir} and DA-CLIP \cite{luo2023controlling} enhance restoration via lightweight prompts and semantic guidance. Recent works such as VLU-Net \cite{zeng2025vision}, DFPIR \cite{tian2025degradation}, and AdaIR \cite{cui2024adair} have pushed forward the progress of AiOIR by incorporating visual-language models, frequency-aware perception, and adaptive strategies.

With the success of diffusion models in image generation, some studies have explored their application in restoration. One group retrains diffusion models for multi-degradation tasks (e.g., WeatherDiffusion \cite{ozdenizci2023restoring}, JCDM \cite{yue2025joint}, DA2Diff \cite{xiong2025da2diff}, DiffUIR \cite{zheng2024selective}), but these approaches are costly and underutilize pre-trained priors. Others fine-tune large models like SD (e.g., AutoDIR \cite{jiang2024autodir}, MPerceiver \cite{ai2024multimodal}, Diff-Restorer \cite{zhang2024diff}), which leverage strong priors but often involve heavy parameters and slow inference. To address these issues, we propose a novel AiOIR method based on SD, combining LoRA for efficient fine-tuning with a one-step diffusion mechanism to significantly accelerate inference while maintaining generation quality.

\begin{figure*}[!t]
    \centering
    \includegraphics[width=1\linewidth]{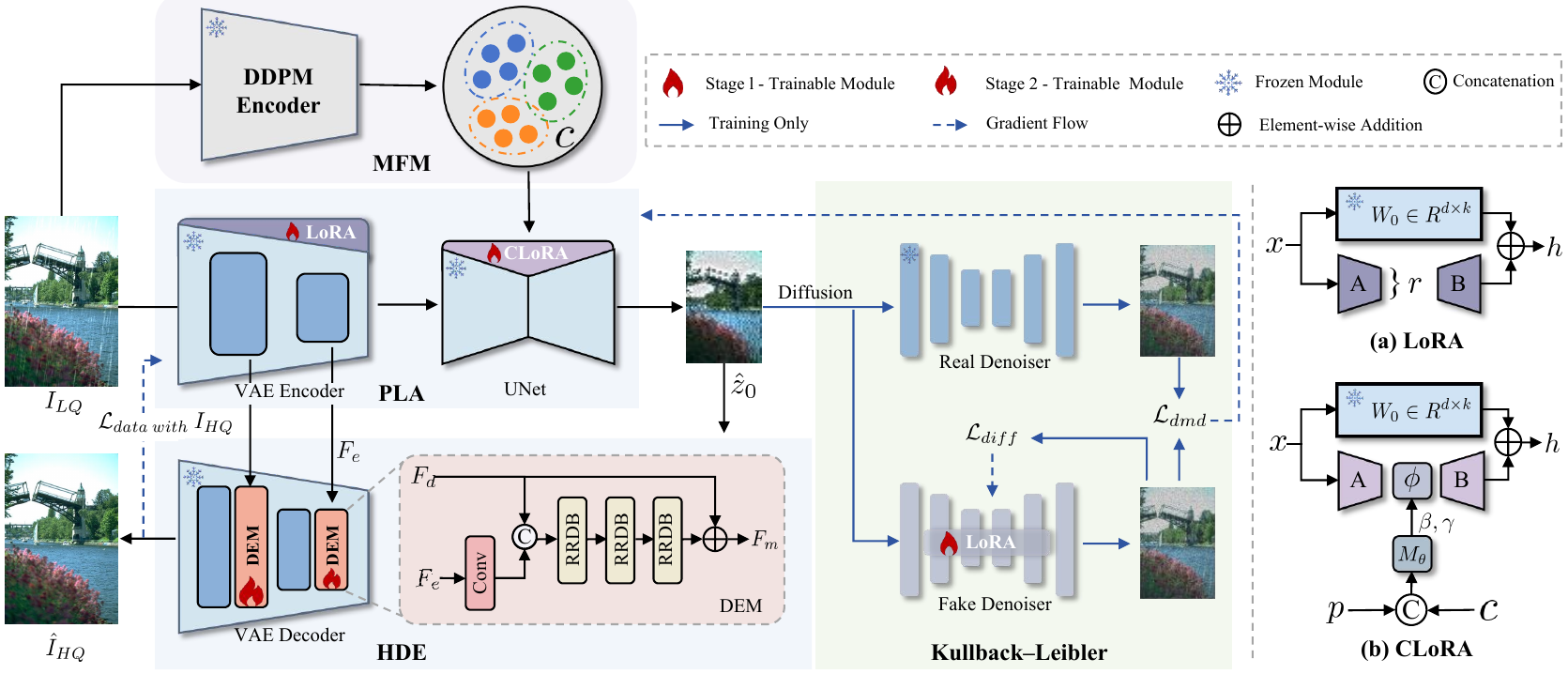}
    \caption{Overview of DOD. DOD enhances SD for AiOIR by injecting LoRA layers into the VAE encoder and conditional LoRA into the UNet. A detail enhancement module is also inserted into the VAE decoder to refine visual details. 
    On the right, (a) shows standard LoRA, while (b) illustrates our proposed conditional LoRA.}
    \label{fig:overview_framework}
\end{figure*}

\subsection{Fine-tuning Strategies for AiOIR}
Rapid advances in large generative models like diffusion models have made leveraging their priors vital for image restoration. Existing methods fall into two categories: structure-guided approaches using ControlNet \cite{zhang2023adding} and lightweight fine-tuning with LoRA \cite{hu2022lora}. ControlNet injects structural priors (e.g., edges, depth) into diffusion models for conditional generation and is widely used in restoration. Key examples include DiffBIR \cite{lin2024diffbir}, AdaptBIR \cite{liu2024adaptbir}, and Diff-Restorer \cite{zhang2024diff}, which improve structural awareness and adaptability. However, these methods often have complex designs and high inference costs.

In contrast, LoRA offers a more efficient and flexible alternative by applying low-rank decomposition to the weight matrices of pre-trained models, requiring only a small number of parameters to be fine-tuned. Methods such as LoRA-IR \cite{ai2024lora} and UIR-LoRA \cite{zhang2024uir} have explored LoRA for multi-degradation modeling, but most are built on CNN or Transformer backbones, with limited integration into diffusion-based frameworks. Moreover, they often rely on stacked modules, leading to structural redundancy and slower response in complex degradation scenarios. To address these challenges, this paper presents an AiOIR method that, for the first time, deeply integrates LoRA’s lightweight adaptation mechanism with the SD model.

\section{Proposed Method}

The DOD model is built upon the SD model, leveraging its exceptional image generation capabilities. To restore images degraded by various types, our model integrates three core modules: Multi-degradation Feature Modulation (MFM), Parameter-efficient conditional LoRA (PLA), and High-fidelity Detail Enhancement (HDE). These modules are designed to flexibly adjust the diffusion priors to accommodate different forms of image degradation.

As shown in Fig.~\ref{fig:overview_framework}, the overall architecture of the model is designed as follows: The low-quality degraded image $I_{LQ}$ is first input into the MFM module to extract degradation feature information $c$. Then, the extracted degradation features, along with the degraded image, are fed into the PLA module. The goal is to effectively remove the degradation with the aid of degradation information, thereby obtaining the potentially restored image features $\hat{z}_0$. Subsequently, these restored latent features are passed into the HDE module to enhance the structural fidelity and texture quality further, ultimately producing a clear and high-quality restored image $\hat{I}_{HQ}$.

\begin{figure}[!t] 
    \centering
    \includegraphics[width=1\linewidth]{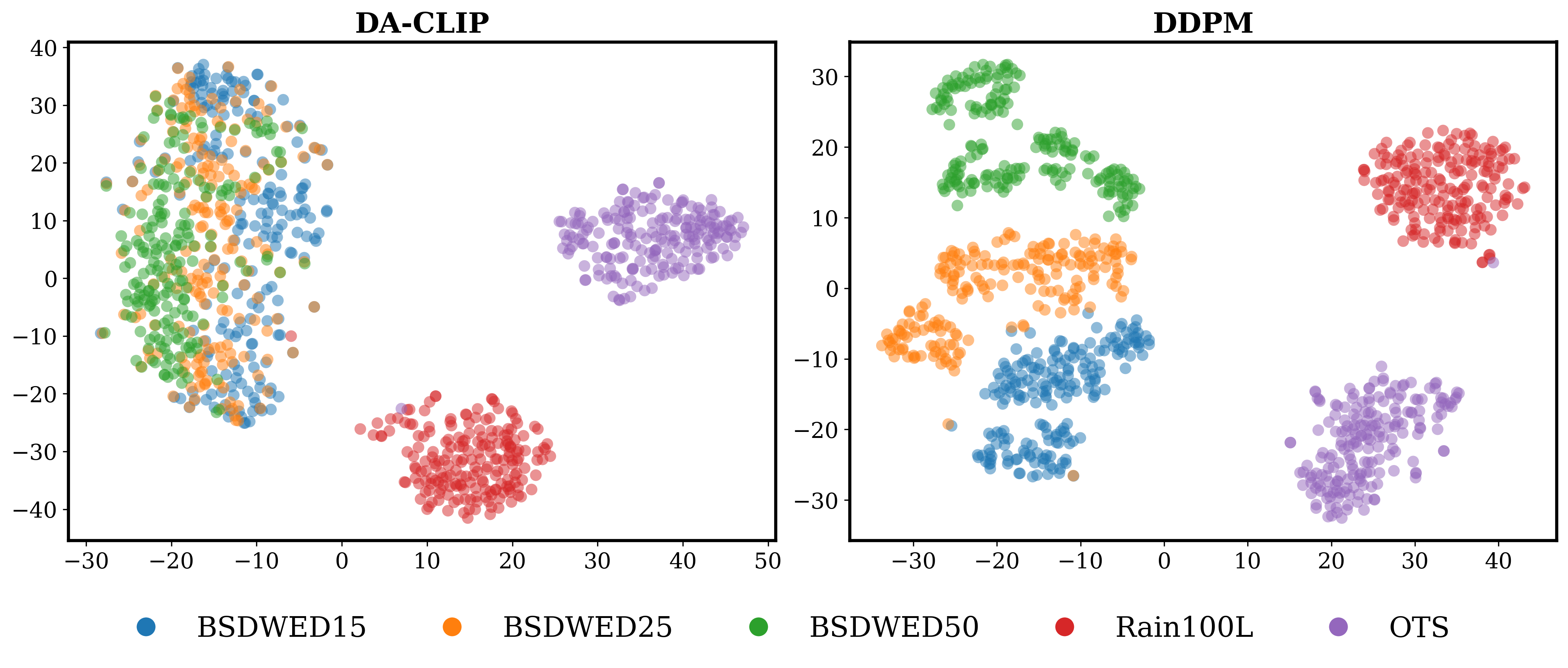} 
    \caption{
    t-SNE visualization of degradation features extracted by DA-CLIP and DDPM.
    DA-CLIP learns degradation-aware prompts by fine-tuning CLIP, while DDPM features are taken from intermediate layers of a pretrained DDPM.
    Compared to DA-CLIP, DDPM features provide clearer separation across degradation types and noise levels, indicating stronger discriminative capacity.
    }
  \label{fig:tsne_comparison}
\end{figure}

\subsection{Multi-degradation Feature Modulation}
Recent studies suggest that recognizing degradation types is crucial for improving both the generalization and performance of image restoration models \cite{hu2025universal}. To enable rapid adaptation to various tasks, we introduce a degradation-aware module that enhances the SD model’s perception of input degradations.

We focus on the h-space, which refers to the intermediate feature space of the UNet within pre-trained DDPM \cite{ho2020denoising}. This space has been shown to contain rich semantic information and offer strong controllability in high-level vision tasks \cite{kwon2022diffusion, jeong2024training}. More recent work, such as DiffL2ID \cite{yang2024unleashing}, further reveals that h-space can also encode low-level degradation features in the context of image dehazing.

Building on these findings, we conduct further experiments and observe that at the diffusion timestep $t=0$, the features in the h-space are clearly separable across different degradation types, as illustrated by the t-SNE visualization in Fig.~\ref{fig:tsne_comparison}. Motivated by this observation, we directly leverage the intermediate features from the pre-trained DDPM as the degradation-aware representation $c$, avoiding additional module training and enabling efficient and generalizable degradation modeling.

\subsection{Parameter-efficient Conditional LoRA}
To enable degradation-aware restoration with minimal parameter overhead, we introduce PLA that leverages LoRA to adapt both the VAE encoder and the UNet in SD, ensuring the generation of high-quality, restored posterior features.
In contrast to existing methods \cite{zhang2024uir, ai2024lora} that typically train multiple LoRA modules independently and combine them based on degradation similarity, we propose a conditional LoRA (CLoRA) mechanism to fine-tune the UNet in SD. By embedding degradation information extracted from MFM into the LoRA structure, CLoRA enables feature space refinement, allowing a single set of parameters to flexibly handle diverse degradation types.


To lay the foundation for our conditional design, we first revisit the standard LoRA formulation as illustrated in Fig.~\ref{fig:overview_framework}(a). Standard LoRA fine-tuning freezes the original weight matrix $W_0 \in \mathbb{R}^{d \times k}$ and introduces a pair of low-rank matrices $A \in \mathbb{R}^{r \times k}$ and $B \in \mathbb{R}^{d \times r}$ to form the adapted weight:
\begin{equation}
W_0 + \Delta W = W_0 + BA,
\end{equation}
where the rank $r$ is chosen such that $r \ll \min(d, k)$. The forward pass of this layer becomes:
\begin{equation}
h = W_0 x + \Delta W x = W_0 x + BAx,
\end{equation}
Typically, $B$ is initialized to zero so that LoRA does not affect the model behavior at the start of fine-tuning.

Based on this standard formulation, we introduce a conditional control mechanism that modulates the low-rank embeddings via a transformation function $\phi(Ax \mid c)$, where $c$ is the degradation information encoded by the MFM module. The overall structure of this CLoRA  is illustrated in Fig.~\ref{fig:overview_framework}(b). Specifically, the output of the CLoRA layer is formulated as:
\begin{equation}
    h = W_0 x + \Delta W x = W_0 x + B \, \phi(Ax \mid c).
\end{equation}

Here, the transformation function $\phi(Ax \mid c)$ adopts a simple affine form:
\begin{equation}
    \phi(Ax \mid c) = \gamma_c \odot Ax + \beta_c,
\end{equation}
where $\odot$ denotes the element-wise multiplication, and the affine parameters $\gamma_c$ and $\beta_c$ are learned from the degradation condition $c$. This CLoRA module offers two key advantages: it enables dynamic, data-dependent parameter adjustment based on the specific degradation characteristics of the low-quality input, and it operates without requiring any manual configuration or user-provided degradation labels.

However, sharing the same affine parameters across all LoRA layers limits the flexibility of fine-tuning. Conversely, if assigning a different affine parameter for each LoRA layer, it causes a significant increase in the total number of trainable parameters due to the construction of a dedicated MLP per layer. To enable flexible parameterization of CLoRA in SD comprising $N$ layers, we introduce a set of learnable prompts $\mathcal{P} = \{p_i\}_{i=1}^N$. The affine parameters for the $i$-th layer, $\gamma_i$ and $\beta_i$, are generated as:
\begin{equation}
    \gamma_i, \beta_i = M_\theta([c, p_i]),
\end{equation}
where $[\cdot, \cdot]$ denotes a concatenation operation, $M_\theta$ is a lightweight network composed of several MLP and normalization layers. 
Compared to methods such as ControlNet~\cite{zhang2023adding}, which duplicate the entire UNet encoder, our approach is superior in both parameter efficiency and computational cost.

\subsection{High-fidelity Detail Enhancement}

To balance inference efficiency and restoration quality, we replace the conventional multi-step sampling in diffusion models with a one-step strategy, which substantially reduces computational overhead but may underperform in recovering fine details. To compensate for this limitation, we introduce a detail enhancement module (DEM) in the VAE decoder to further refine textures and enhance image fidelity.

Specifically, features extracted by the encoder are first passed through a convolutional layer initialized with zeros to align their dimensions with the decoder features. These aligned features are then concatenated with the corresponding decoder features and fed into a series of residual-in-residual dense blocks (RRDB)  \cite{wang2018esrgan} to extract detail-aware representations. The output is subsequently fused with the original decoder features to compensate for the detail loss caused by one-step diffusion. Additionally, a learnable dynamic weight $w$ is introduced to control the refinement strength. The overall process can be formulated as:
\begin{equation}
F_m = F_d + \mathrm{RRDB}([\mathrm{Conv}(F_e), F_d]; \theta) \times w,
\end{equation}
where $F_e$ and $F_d$ denote the features from the VAE encoder and decoder, respectively, and $F_m$ is the final fused feature. $\mathrm{Conv}(\cdot)$ represents a convolutional layer initialized with zeros and $\mathrm{RRDB}(\cdot\,; \theta)$ denotes the RRDB module with learnable parameters $\theta$. 

\subsection{Optimization}
To stabilize the training process, we adopt a two-stage strategy to optimize the whole model.

\noindent\textbf{Stage 1} aims to efficiently adapt the SD model for image restoration tasks. In this stage, we train only the PLA module while keeping the VAE decoder frozen, decoding the generated posterior feature $\hat{z}_0$ to obtain the restored image $\hat{I}_{HQ}$. During training, standard data consistency losses are employed to measure the error of alignment between the reconstructed and clean images at both pixel and semantic levels. To enable one-step diffusion, degradation-aware image restoration while preserving the generative capacity of the pretrained model, we further incorporate the distribution matching distillation (DMD) loss \cite{yin2024one}, which enhances both generative quality and generalization ability.


\textit{ 1) Data consistency loss.}
The data consistency loss $\mathcal{L}_{\text{data}}$ consists of a pixel-wise reconstruction term and a perceptual similarity term, formulated as:
\begin{equation}
\mathcal{L}_{\text{data}} = \lambda_{\text{rec}} \cdot \mathcal{L}_{\text{rec}} + \lambda_{\text{lpips}} \cdot \mathcal{L}_{\text{lpips}},
\end{equation}
\begin{equation}
\mathcal{L}_{\text{rec}} = \| \hat{I}_{HQ} - I_{HQ} \|_2,
\end{equation}
\begin{equation}
\mathcal{L}_{\text{lpips}} = \| \Phi(\hat{I}_{HQ}) - \Phi(I_{HQ}) \|_2,
\end{equation}
where $\hat{I}_{HQ}$ denotes the generated image and $I_{HQ}$ is the ground-truth clean image. $\Phi(\cdot)$ represents perceptual features extracted by VGG \cite{simonyan2014very} network. $\lambda_{\text{rec}}$ and $\lambda_{\text{lpips}}$ balance the two loss terms.

\textit{ 2) Distribution matching distillation loss.}
To enable the network to rapidly generate samples that are indistinguishable from real images, the loss function \cite{yin2024one} minimizes the KL divergence between the real image feature distribution $p_{\text{real}}$ and the generated image feature distribution $p_{\text{fake}}$.
\begin{equation}
\mathcal{L}_{\text{dmd}} = \text{KL}\left( p_{\text{real}} \parallel p_{\text{fake}} \right).
\end{equation}

\begin{table*}[t]
\centering
\small
\setlength{\tabcolsep}{4pt} 
\renewcommand{\arraystretch}{1.2} 

\begin{tabular}{clccccccccc}
\toprule
& \textbf{Method} & \textbf{Venue} 
& \textbf{PSNR$\uparrow$} & \textbf{SSIM$\uparrow$} 
& \textbf{LPIPS$\downarrow$} & \textbf{DISTS$\downarrow$} 
& \textbf{CLIPIQA$\uparrow$} & \textbf{NIQE$\downarrow$} 
& \textbf{MUSIQ$\uparrow$} & \textbf{MANIQA$\uparrow$} \\
\midrule
\multirow{6}{*}{\rotatebox[origin=c]{90}{\textbf{Non-Diff}}}
& AirNet    & CVPR2022  & 31.11 & 0.9068 & 0.0925 & 0.0950 & 0.6405 & 3.3911 & 66.98 & 0.6470 \\
& PromptIR  & NeurIPS23 & 32.18 & 0.9124 & 0.0859 & 0.0864 & 0.6409 & 4.0061 & 67.21 & 0.6569 \\
& LORA-IR   & arXiv2024 & 30.98 & 0.9031 & 0.0889 & 0.0809 & 0.6352 & 3.4133 & 67.22 & 0.6568 \\
& VLU-Net   & CVPR2025  & 32.55 & 0.9157 & 0.0796 & 0.0785 & 0.6433 & 3.6559 & 67.09 & 0.6698 \\
& DFPIR     & CVPR2025  & 32.75 & \textbf{0.9162} & 0.0758 & 0.0758 & 0.6626 & 3.5938 & 67.34 & 0.6679 \\
& AdaIR     & ICLR2025  & \textbf{32.98} & 0.9155 & 0.0825 & 0.0820 & 0.6435 & 3.6464 & 67.50 & 0.6685 \\
\midrule
\multirow{3}{*}{\rotatebox[origin=c]{90}{\textbf{Diff}}}
& DA-CLIP   & ICLR2024  & 30.27 & 0.8780 & 0.0664 & 0.0650 & 0.6580 & \textbf{3.2783} & 67.31 & 0.6663 \\
& DiffUIR   & CVPR2024  & 31.89 & 0.9010 & 0.0959 & 0.0964 & 0.6163 & 3.7371 & 67.51 & 0.6570 \\
& Ours & -     & 31.87 & 0.9122 & \textbf{0.0620} & \textbf{0.0626} & \textbf{0.6660} & 3.5212 & \textbf{68.85} & \textbf{0.7080} \\
\bottomrule
\end{tabular}

\caption{Quantitative comparison on average performance across three restoration tasks: dehazing, deraining, and Gaussian denoising ($\sigma = 15$, $25$, $50$). The best results are marked in boldface.}
\label{tab:comparison}
\end{table*}
In practice, we directly backpropagate through the KL divergence, whose gradient is computed by estimating the scores of the real and generated distributions. To this end, we employ a pair of diffusion denoisers to model the score functions of real and fake distributions after Gaussian diffusion. As illustrated in Fig. \ref{fig:overview_framework}, both denoisers adopt the UNet architecture from SD and are initialized with the same pretrained weights. The real denoiser keeps its parameters frozen throughout training, while the fake denoiser is dynamically updated according to the evolving distribution of generated samples. It is jointly optimized with the image restoration network using the standard noise prediction loss $\mathcal{L}_{\text{diff}}$ from SD. Please refer to the supplementary material for the detailed loss.


The overall loss for stage 1 of DOD formulated as:
\begin{equation}
\mathcal{L}_{\text{stage1}} = \mathcal{L}_{\text{data}} + 
\mathcal{L}_{\text{dmd}}.
\end{equation}

\noindent\textbf{Stage 2} focuses on recovering the missing fine details from the latent features reconstructed in the previous stage. To achieve this, the PLA module fine-tuned in Stage 1 is frozen, and only the DEMs, which are inserted into each layer of the VAE decoder as part of the HDE, are trained. A combination of reconstruction loss and structural similarity loss is employed to enhance perceptual quality and detail fidelity.

\textit{Structural similarity loss.}
To promote structural integrity and contrast preservation, we use SSIM-based loss \cite{wang2004image}:
\begin{equation}
\mathcal{L}_{\text{ssim}} = 1 - \text{SSIM}(\hat{I}_{HQ}, I_{HQ}),
\end{equation}
which penalizes discrepancies in luminance, contrast, and structural patterns between $\hat{I}_{HQ}$ and $I_{HQ}$.


The total loss at stage 2 is:
\begin{equation}
\mathcal{L}_{\text{stage2}} = \lambda_{\text{rec}} \cdot \mathcal{L}_{\text{rec}} + 
\lambda_{\text{ssim}} \cdot \mathcal{L}_{\text{ssim}},
\end{equation}
where $\lambda_{\text{rec}}$ and $\lambda_{\text{ssim}}$ are hyperparameters that balance each component's contribution.

\section{Experiments}
\subsection{Experimental Settings}


\textbf{Datasets.} Following previous works \cite{tian2025degradation, li2022all}, we construct task-specific datasets and evaluate our method under two standard AiOIR settings. The first setting, rain-haze-noise, includes five degradation types across multiple datasets: BSD400 \cite{arbelaez2010contour} and WED \cite{ma2016waterloo} for denoising (tested on BSD68 \cite{martin2001database}), Rain100L \cite{yang2017deep} for deraining, and RESIDE \cite{li2018benchmarking} (OTS for training, SOTS for testing) for dehazing. The second setting, rain-haze-noise-blur-dark, adds deblurring and low-light enhancement, using GoPro \cite{nah2017deep} and LOL \cite{wei2018deep} datasets respectively. 

\noindent\textbf{Metrics.}
To comprehensively evaluate the performance of image restoration methods, we utilize multiple image quality assessment metrics.
Specifically, PSNR and SSIM \cite{wang2004image} are full-reference metrics that measure image fidelity, while LPIPS \cite{zhang2018unreasonable} and DISTS \cite{ding2020image} assess perceptual quality. In addition, NIQE \cite{zhang2015feature}, MANIQA \cite{yang2022maniqa}, MUSIQ \cite{ke2021musiq}, and CLIPIQA \cite{wang2023exploring} are no-reference metrics that enable quality evaluation without access to ground-truth (GT) images.

\subsection{Comparison with State-of-the-Arts}
\subsubsection{Compared methods.} We benchmark our method against nine state-of-the-art AiOIR baselines: AirNet \cite{li2022all}, PromptIR  \cite{potlapalli2023promptir}, LORA-IR \cite{ai2024lora}, DA-CLIP \cite{luo2023controlling}, DiffUIR \cite{zheng2024selective}, VLU-Net  \cite{zeng2025vision}, DFPIR \cite{tian2025degradation}, AdaIR  \cite{cui2024adair}, and AutoDIR \cite{jiang2024autodir}. Among them, DA-CLIP and DiffUIR train diffusion models from scratch in the pixel domain, whereas AutoDIR is built upon the pre-trained SD. The remaining approaches follow conventional, non-generative paradigms.
\begin{figure*}[!t] 
    \centering
    \includegraphics[width=\textwidth]{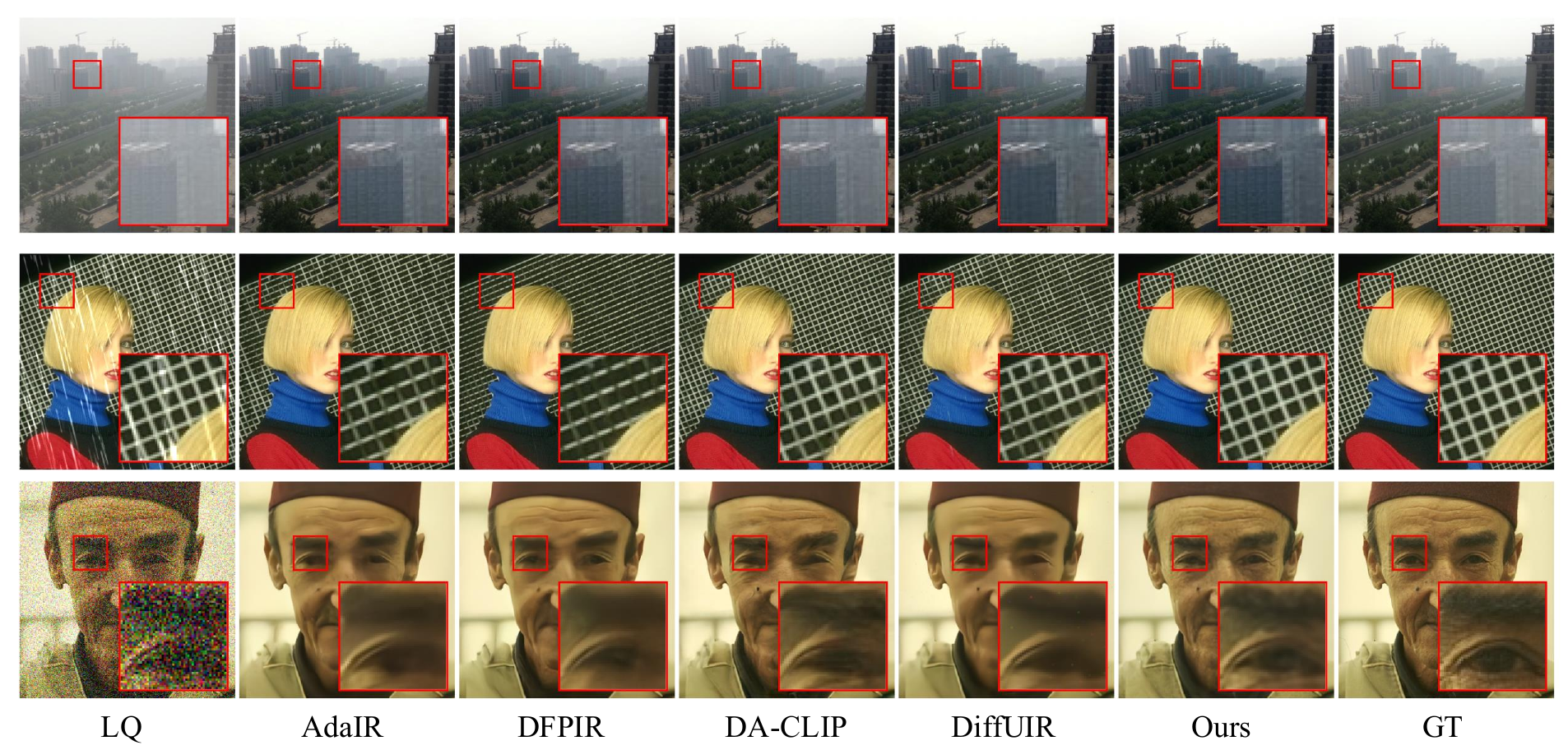} 
    \caption{
    Visual comparisons with state-of-the-art AiOIR methods on three tasks. Please zoom in for a better view. 
    }
  \label{fig:visual_compare}
\end{figure*}

\begin{table*}[t]
\centering
\small
\setlength{\tabcolsep}{2pt} 
\begin{tabular}{clcccccccccc}
\toprule
\multirow{2}{*}{} & \multirow{2}{*}{\textbf{Method}} 
& \multicolumn{2}{c}{\textbf{Dehazing}} 
& \multicolumn{2}{c}{\textbf{Deraining}} 
& \multicolumn{2}{c}{\textbf{Denoising}} 
& \multicolumn{2}{c}{\textbf{Deblurring}} 
& \multicolumn{2}{c}{\textbf{Low-light}} \\
& & MUSIQ$\uparrow$ & MANIQA$\uparrow$ & MUSIQ$\uparrow$ & MANIQA$\uparrow$ & MUSIQ$\uparrow$ & MANIQA$\uparrow$ & MUSIQ$\uparrow$ & MANIQA$\uparrow$ & MUSIQ$\uparrow$ & MANIQA$\uparrow$ \\
\midrule
\multirow{4}{*}{\rotatebox[origin=c]{90}{\small\textbf{Non-Diff}}}
& PromptIR     & 65.43 & 0.6829 & 68.00 & 0.6679 & 67.29 & 0.6295 & 36.60 & 0.4250 & 64.76 & 0.5922 \\
& AdaIR        & 65.83 & 0.6923 & 68.31 & 0.6720 & 67.65 & 0.6593 & 33.41 & 0.3763 & 69.43 & 0.6309 \\
& VLU-Net      & 65.81 & 0.6920 & 68.45 & 0.6740 & 68.38 & 0.6669 & 32.48 & 0.3568 & 65.21 & 0.5948 \\
& DFPIR        & 65.61 & 0.6898 & 68.29 & 0.6700 & 68.06 & 0.6563 & 36.93 & 0.3978 & 69.19 & 0.6234 \\
\midrule
\multirow{3}{*}{\rotatebox[origin=c]{90}{\small\textbf{Diff}}}
& DA-CLIP      & 67.88 & 0.6524 & 68.60 & 0.6695 & 67.88 & 0.6524 & 36.89 & \textbf{0.4298} & 72.59 & 0.6619 \\
& DiffUIR      & 64.84 & 0.6864 & 68.40 & 0.6682 & 68.19 & 0.6383 & 33.63 & 0.3529 & 71.13 & 0.6246 \\

& Ours 
& \textbf{67.99} & \textbf{0.7115} & \textbf{70.49} & \textbf{0.7104} & \textbf{69.50} & \textbf{0.7219} & \textbf{37.77} & 0.4286 & \textbf{73.75} & \textbf{0.6657}\\
\bottomrule
\end{tabular}
\caption{
Quantitative comparison of no-reference metrics MUSIQ and MANIQA across five restoration tasks: dehazing, deraining, denoising ($\sigma = 25$), deblurring, and low-light enhancement. The best results are marked in boldface.
}
\label{tab:five-task}
\end{table*}
\subsubsection{Quantitative comparisons.} 
Table~\ref{tab:comparison} presents the average quantitative results of our method across three image restoration tasks. It is observed that non-diffusion methods perform strongly on traditional fidelity metrics such as PSNR and SSIM, with AdaIR and DFPIR achieving the highest scores of 32.98 dB (PSNR) and 0.9162 (SSIM), respectively. However, these methods fall short on perceptual quality metrics, making it challenging to satisfy the demands of real-world visual experience.

In contrast, diffusion-based methods, exemplified by DA-CLIP and our approach, show clear advantages in perceptual metrics. Their restoration results better align with human visual preferences in terms of detail preservation and structural integrity. Notably, our method achieves the best scores across five perceptual metrics, demonstrating superior perceptual quality. Moreover, while maintaining perceptual performance comparable to DA-CLIP, our method significantly improves fidelity, with PSNR and SSIM increasing by 1.60 dB and 0.034 respectively, further validating its strengths in restoring image details and preserving structure.

Furthermore, we extend the evaluation to a more challenging five-task setting by incorporating deblurring and low-light enhancement. Table~\ref{tab:five-task} summarizes the performance of each task under two perceptual quality metrics, MUSIQ and MANIQA. Our method consistently achieves the best or second-best results across all tasks, with average improvements of approximately 3\% to 6\% on MUSIQ and 3\% to 8\% on MANIQA, highlighting strong generalization capability. Particularly in perceptually demanding tasks such as deblurring and low-light enhancement, our method leads in perceptual scores, further confirming its effectiveness in restoring details across diverse degradation types. These results demonstrate that our unified framework not only excels in the three-task setting but also exhibits high adaptability and robustness in more complex multi-task scenarios.

\subsubsection{Qualitative comparisons.}
Fig. \ref{fig:visual_compare} presents a visual comparison between our method and representative approaches across three tasks. Although existing methods can reduce image degradation to some extent, the enlarged regions highlight their deficiencies in detail restoration. For instance, in the denoising task, while various methods successfully remove noise, the resulting images are overly smooth and lack detail, negatively impacting visual quality. In contrast, our method not only effectively removes noise but also better preserves fine details such as facial textures. This demonstrates the superior stability and performance of our method across multiple image restoration tasks. Additional visual comparisons can be found in the supplementary material.

\subsubsection{Inference speed comparison.}
We compare four representative diffusion-based restoration methods in terms of inference steps and time. In Table \ref{tab:inference_comparison}, our method achieves the fastest inference with just one step. Although WeatherDiff uses only 25 steps, its patch-based overlapping strategy makes it much slower. Compared to DA-CLIP, which offers similar visual quality, our method is about 84× faster, demonstrating clear efficiency advantages.

\begin{table}[htbp]
\centering
\small  
\setlength{\tabcolsep}{1.5pt}  
\begin{tabular}{cccccc}
\toprule
\textbf{Method} & WeatherDiff & AutoDIR & DA-CLIP & DiffUIR & Ours \\
\midrule
\textbf{Step $\downarrow$}     & 25      & 200     & 100     & 3      & \textbf{1}     \\
\textbf{Time(s) $\downarrow$} & 212.534 & 26.602  & 17.664  & 0.303  & \textbf{0.211} \\
\bottomrule
\end{tabular}
\caption{Comparison of inference steps and time. All methods are tested with an input image of size $512 \times 512$, and the inference time is measured on a 3090 GPU. The best results are marked in boldface.}
\label{tab:inference_comparison}
\end{table}

\subsection{Ablation Study}
\begin{figure}[htbp] 
    \centering
    \includegraphics[width=1\linewidth]{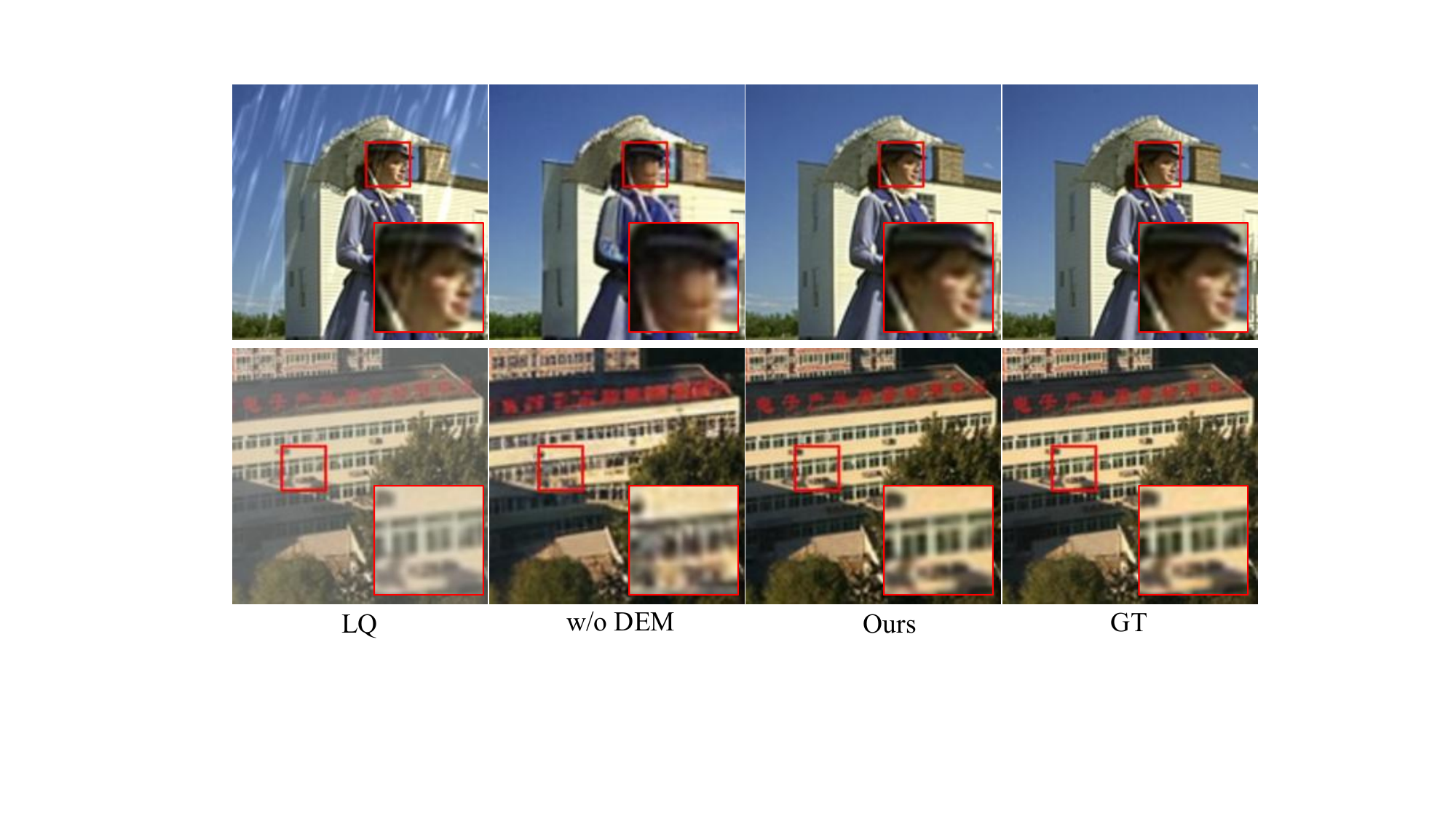} 
    \caption{
    With DEM, fine structures are well preserved; without it, degradation is removed but key details are lost.
    }
  \label{fig:ablation}
\end{figure}

\noindent\textbf{Effectiveness of CLoRA and DEM.}
The results in Table~\ref{tab:ablation_study} demonstrate that CLoRA and DEM play complementary roles in image restoration. Introducing conditional LoRA improves MUSIQ by 2.19 and reduces DISTS by 22\%, indicating a positive impact on structural consistency and perceptual quality. DEM, on the other hand, focuses on enhancing fine details. When applied independently, it increases PSNR by 8.78 dB and MUSIQ by 3.49, highlighting its effectiveness in reconstructing low-level features. This is further illustrated in the visual comparisons shown in Fig.~\ref{fig:ablation}, where DEM successfully restores fine-grained textures of the original image. When combined, the two modules achieve the best performance across PSNR, DISTS, and MUSIQ, suggesting a strong synergistic effect. These findings confirm that the proposed design effectively balances fidelity and perceptual quality, and validate its applicability to multi-task image restoration.

\begin{table}[htbp]
\centering
\small
\setlength{\tabcolsep}{4pt}  
\begin{tabular}{cccccc}
\toprule
\textbf{CLoRA} & \textbf{DEM} & \textbf{PSNR$\uparrow$} & \textbf{DISTS$\downarrow$} & \textbf{NIQE$\downarrow$} & \textbf{MUSIQ$\uparrow$} \\
\midrule
\ding{55} & \ding{55} & 23.05 & 0.1121 & 4.4220 & 64.81 \\
\ding{51} & \ding{55} & 24.60 & 0.0873 & 3.7072 & 67.00 \\
\ding{55} & \ding{51} & 31.83 & 0.0666 & 3.5118 & 68.30 \\
\ding{51} & \ding{51} & 31.87 & 0.0626 & 3.5212 & 68.85 \\
\bottomrule
\end{tabular}
\caption{Ablation study of CLoRA and DEM.}
\label{tab:ablation_study}
\end{table}

\begin{table}[htbp]
\centering
\small
\setlength{\tabcolsep}{2.2pt}  
\begin{tabular}{ccccc}
\toprule
\textbf{Method} & \textbf{PSNR$\uparrow$} & \textbf{DISTS$\downarrow$} & \textbf{NIQE$\downarrow$} & \textbf{MUSIQ$\uparrow$} \\
\midrule
VAE Encoder + UNet & 30.68 & 0.0666 & 3.4106 & 68.58 \\
UNet only          & 31.87 & 0.0626 & 3.5212 & 68.85 \\
\bottomrule
\end{tabular}
\caption{Performance comparison of CLoRA placement.}
\label{tab:condlora_placement}
\end{table}
\noindent\textbf{Effect of CLoRA placement.} As shown in Table \ref{tab:condlora_placement}, applying it solely to the UNet leads to the best performance, achieving a PSNR of 31.87 dB with corresponding improvements in DISTS and MUSIQ. This suggests that CLoRA is more effective when used to modulate the generation process closely related to degradation. In contrast, introducing conditional modulation into the VAE encoder may disrupt its representation ability and negatively impact overall restoration quality.

\noindent\textbf{The number of LoRA rank.}
As shown in Table~\ref{tab:lora_rank}, we evaluate the impact of different LoRA ranks across three tasks when jointly fine-tuning the VAE encoder and UNet. A small rank $r=4$ leads to the model collapse directly, while a large rank $r=16$ brings a certain degree of performance drop. In contrast, a rank of 8 achieves the best balance between stability and performance. Thus, we set the LoRA rank to 8 for both the fine-tuning of VAE encoder and UNet.

\begin{table}[htbp]
\centering
\small
\setlength{\tabcolsep}{6pt}  
\begin{tabular}{ccccc}
\toprule
\textbf{Rank} & \textbf{PSNR$\uparrow$} & \textbf{DISTS$\downarrow$} & \textbf{NIQE$\downarrow$} & \textbf{MUSIQ$\uparrow$} \\
\midrule
4  & --    & --     & --     & --    \\
8  & 31.87 & 0.0620 & 3.5212 & 68.85 \\
16 & 31.68 & 0.0626 & 3.5409 & 68.13 \\
\bottomrule
\end{tabular}
\caption{Comparison of LoRA in VAE encoder and UNet with different ranks.}
\label{tab:lora_rank}
\end{table}

\section{Conclusion}
This paper presents an efficient all-in-one image restoration method, dubbed Diffusion Once and Done (DOD). By incorporating a degradation-aware LoRA fine-tuning strategy and a two-stage structural optimization mechanism, the method enables image restoration with various degradation types. 
Combined with variational fractional distillation, it produces high-quality restored images using one-step sampling while significantly reducing fine-tuning costs. 
Experimental results demonstrate that the proposed method achieves strong performance in subjective quality and objective fidelity across multiple restoration tasks.
Despite the promising performance of one-step diffusion in image restoration tasks, the computational cost remains a limiting factor for deployment on resource-constrained devices. In future work, we plan to explore model compression techniques such as quantization to further reduce inference overhead. 

\bibliography{aaai2026}

\end{document}